\documentclass[sigconf]{acmart}

\newcommand{\etal}{\textit{et al.}}
\newcommand{\ie}{\textit{i.e.}}
\newcommand{\eg}{\textit{e.g.}}

\AtBeginDocument{%
  }

\renewcommand\footnotetextcopyrightpermission[1]{}

\begin{document}

\title{EmoCtrl: Controllable Emotional Image Content Generation}

\author{Jingyuan Yang}
\affiliation{%
  \institution{CSSE, Shenzhen University}
  \state{Shenzhen}
  \country{China}
}
\email{jingyuanyang.jyy@gmail.com}

\author{Weibin Luo}
\affiliation{%
  \institution{CSSE, Shenzhen University}
  \state{Shenzhen}
  \country{China}
}
\email{waibunlok@gmail.com}

\author{Hui Huang}
\authornote{Corresponding author.}
\affiliation{%
  \institution{CSSE, Shenzhen University}
  \state{Shenzhen}
  \country{China}
}
\email{hhzhiyan@gmail.com}

\renewcommand{\shortauthors}{Yang et al.}

\begin{abstract}
    An image conveys meaning through both its visual content and emotional tone, jointly shaping human perception.
    We introduce Controllable Emotional Image Content Generation (C-EICG), which aims to generate images that remain faithful to a given content description while expressing a target emotion.
    Existing text-to-image models ensure content consistency but lack emotional awareness, whereas emotion-driven models generate affective results at the cost of content distortion.
    To address this gap, we propose EmoCtrl, supported by a dataset annotated with content, emotion, and affective prompts, bridging abstract emotions to visual cues.
    EmoCtrl incorporates textual and visual emotion enhancement modules that enrich affective expression via descriptive semantics and perceptual cues.
    To align with human preference, we further introduce an emotion-driven preference optimization with specifically designed emotion reward.
    Comprehensive experiments demonstrate that EmoCtrl achieves faithful content and expressive emotion control, outperforming existing methods.
    User studies confirm EmoCtrl's strong alignment with human preference.
    Moreover, EmoCtrl generalizes well to creative applications, further demonstrating the robustness and adaptability of the learned emotion tokens.
\end{abstract}


\begin{teaserfigure}
  \includegraphics[width=\linewidth]{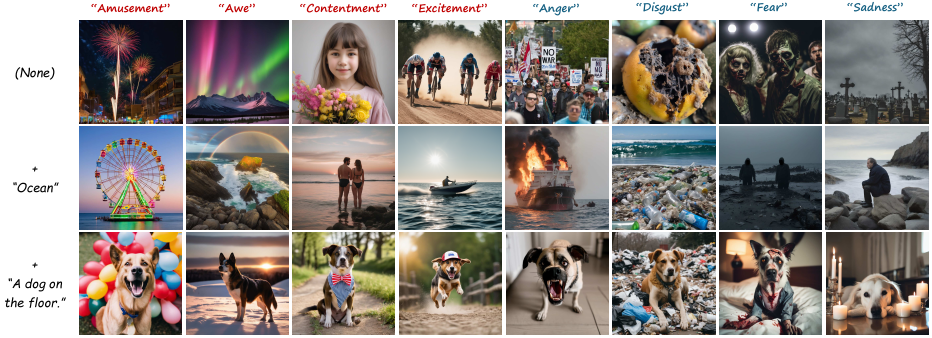}
  \caption{Controllable Emotional Image Content Generation with EmoCtrl. Given a content condition (\textit{``Ocean''}) and a target emotion (\textit{``Contentment''}), EmoCtrl generates images that maintain semantic content while vividly expressing accurate emotions.}
  \Description{}
  \vspace{10pt}
  \label{fig:teaser}
\end{teaserfigure}

\maketitle

\section{Introduction}
\label{sec:intro}

\begin{flushleft}
	\textit{``The artist must train not only his eye but also his soul.''}
\end{flushleft}
\begin{flushright}
	\textit{--Wassily Kandinsky}
\end{flushright}

Images are shaped not only by \textit{what} they show, but also by \textit{how} they make us feel.
Understanding and manipulating these affective cues is central to many applications, from artistic creation~\cite{giorgescu2024affective, loren2025creation} to human-computer interaction~\cite{song2025affective, gu2025review}.
The same content can evoke different emotions in different contexts, and each emotion can be triggered by diverse visual cues~\cite{yang2021solver}.
For example, a fruitful tree may convey \textit{contentment}, while a withered one evokes \textit{sadness}.
Likewise, a sunny sky may elicit \textit{amusement}, whereas a stormy sky evokes \textit{fear}.
These observations highlight the necessity of controlling content and emotion as distinct dimensions, raising the question of whether images can be generated that are both semantically faithful and emotionally expressive.

We introduce Controllable Emotional Image Content Generation (C-EICG), a task that generates images matching the given content while expressing a target emotion.
The content specifies what to depict, provided as either as a concept (\eg, ``Ocean'') or a caption (\eg, ``A dog on the floor''), as shown in Fig.~\ref{fig:teaser}.
The target emotion is selected from eight Mikels' categories~\cite{mikels2005emotional}: amusement, awe, contentment, excitement, anger, disgust, fear, and sadness.

\begin{figure}
	\centering
	\includegraphics[width=0.95\linewidth]{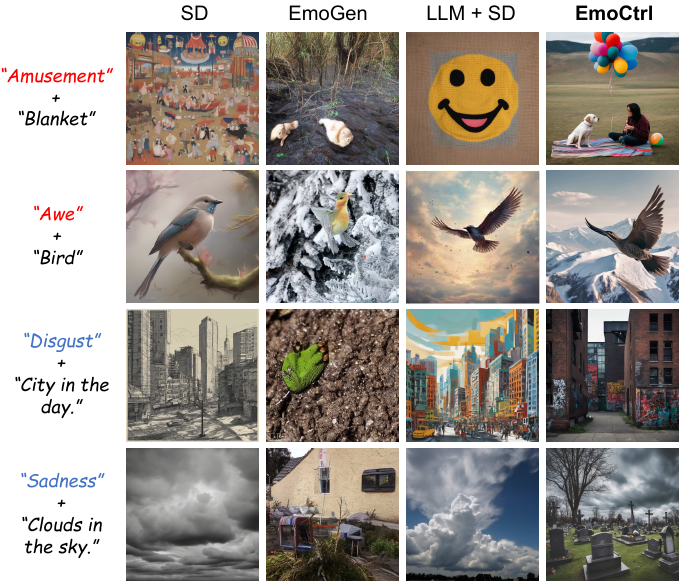}
	\vspace{-5pt}
	\caption{SD lacks emotional expressiveness, EmoGen struggles with content control, and LLM + SD produces ambiguous semantics. In contrast, EmoCtrl generates images that faithfully align with both emotion and content conditions.}
	\Description{}
	\label{fig:intro}
	\vspace{-20pt}
\end{figure}

Naturally, we examine whether existing methods can address this task.
As shown in Fig.~\ref{fig:intro}, neither text-to-image nor emotion-driven models can jointly control content and emotion.
Stable Diffusion (SD)~\cite{rombach2022high} preserves semantics but lacks emotional awareness, while EmoGen~\cite{yang2024emogen} enhances emotional expression at the cost of content preservation.
Although Large Language Models (LLMs)~\cite{achiam2023gpt} may encode emotional knowledge, combining them with SD still falls short.
LLMs describe emotions mainly as abstract concepts (\eg, ``evoking fear'') or simple facial expressions (\eg, happy), revealing their inability to translate abstract emotion words into affective, content-aligned images.

Existing datasets like EmoSet~\cite{yang2023emoset} provide only emotion labels, hindering the joint control of content and emotion.
Meanwhile, text-to-image models require detailed prompts, yet EmoSet provides only short emotion attributes.
To overcome these issues, we build a new dataset featuring both content and emotion annotations, supplemented with affective prompts that bridge abstract emotions and expressive visual representations.
To be specific, based on EmoSet and EmoEditSet~\cite{yang2025emoedit}, we enrich each image with detailed captions and extracted concepts using multimodal large language models (MLLMs)~\cite{bai2025qwen2} and LLMs.
After data filtering and human verification, we obtain EmoSet+ and EmoEditSet+, where each image is annotated with both emotion and content labels, together with detailed affective prompts.

In this work, we present EmoCtrl, a C-EICG framework that enhances emotional expression through two complementary modules: Textual Emotion Enhancement and Visual Emotion Enhancement.
The former enriches affective semantics in text, while the latter reinforces emotional cues in visual domain.
To mitigate the biased interpretation of emotion words in LLMs and SD, we assign a learnable token to each emotion category for each modality.
Textual Emotion Enhancement integrates textual emotion tokens with content descriptions via a LoRA-tuned LLM to produce emotion-aware prompts.
Visual Emotion Enhancement injects visual emotion tokens into diffusion model through cross-attention to improve emotional fidelity while preserving content consistency.
Furthermore, we introduce Emotion-driven Preference Optimization, which incorporates carefully designed emotion and polarity rewards alongside the general PickScore.
This design enhances EmoCtrl not only at textual and visual levels, but also in its alignment with human preferences, thereby improving both emotion accuracy and intensity.

We evaluate EmoCtrl on an inference set comprising 76 content descriptions paired with 8 target emotions.
Across emotion fidelity, content alignment, and joint control, EmoCtrl mostly outperforms existing text-to-image and emotion-driven models.
User study validates EmoCtrl’s strong alignment with human preference, while ablation study demonstrates the effectiveness of its key components.
In addition, EmoCtrl extends to creative tasks, including stylized and multi-emotion C-EICG.

In summary, our contributions are:
\begin{itemize}

	\item We introduce Controllable Emotional Image Content Generation (C-EICG), curate tailored datasets for this task, and propose EmoCtrl, a framework that generates images aligned with both the given content and the target emotion.

	\item We design two complementary emotion enhancement modules, Textual Emotion Enhancement and Visual Emotion Enhancement, and further propose Emotion-driven Preference Optimization to align human preference.

	\item We evaluate EmoCtrl against state-of-the-art methods and demonstrate our superiority in both evoking specific emotion and preserving the given content which is further validated by image-generation applications.

\end{itemize}


\section{Related work}
\label{sec:rw}

\begin{figure*}
    \centering
    \includegraphics[width=\linewidth]{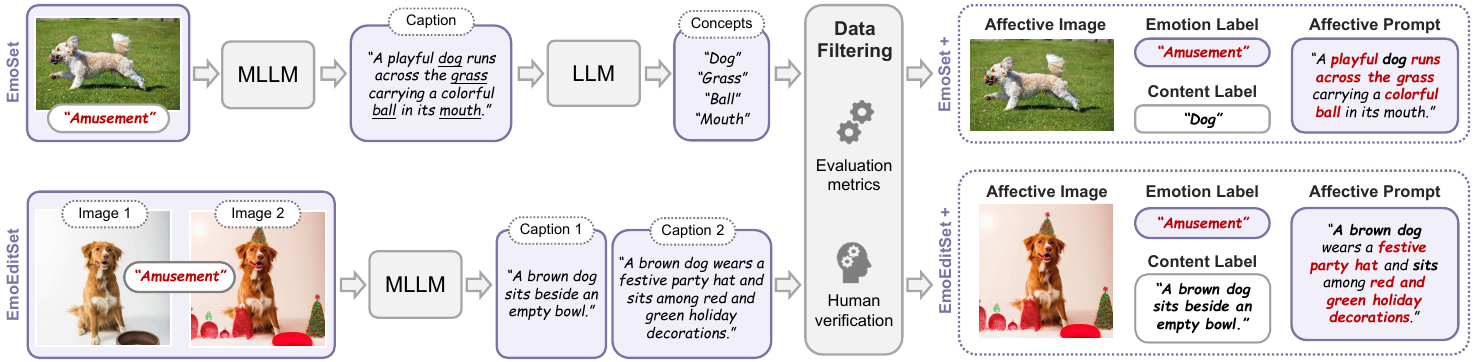}
    \vspace{-10pt}
    \caption{Data construction. EmoSet+ and EmoEditSet+ are constructed with triplet annotations linking emotion, content, and caption, serving as supervision for controllable EICG.}
    \Description{}
    \label{fig:method_1}
    \vspace{-15pt}
\end{figure*}

\subsection{Text-to-Image Generation}
Text-to-image (T2I) generation aims to synthesize images that align with user-provided text prompts~\cite{yang2025text}.

\paragraph{Diffusion-based Text-to-Image Generation.}
Diffusion-based T2I generation has achieved remarkable progress in recent years~\cite{zhang2023text}.
Stable Diffusion (SD)~\cite{rombach2022high} performs latent-space denoising with CLIP-based guidance~\cite{radford2021clip},
while SDXL~\cite{podell2023sdxl} further improves generation fidelity through a larger architecture and multi-stage refinement.
Chen \etal~\cite{chen2023pixart} proposed Pixart-$\alpha$, a transformer-based diffusion model~\cite{peebles2023scalable} achieving superior text–image alignment and training efficiency.
To enhance controllability, ControlNet~\cite{zhang2023adding} introduces additional spatial conditions such as edges and depth maps into the diffusion process, while T2I-Adapter~\cite{mou2024t2i} provides a plug-and-play module for conditioning on various visual modalities.
Ye \etal~\cite{ye2023ip} extended this direction with IP-Adapter, aligning diffusion features with external image embeddings for more flexible cross-modal guidance.
Beyond conditional control, several methods focus on learning new visual concepts.
Textual Inversion~\cite{gal2022ti} learns new token embeddings without modifying model weights, while DreamBooth~\cite{ruiz2023dreambooth} fine-tunes the diffusion model to bind new visual identities to textual tokens.

\paragraph{LLM-Enhanced Text-to-Image Generation.}
Large Language Models (LLMs)~\cite{touvron2023llama, touvron2023llama2, grattafiori2024llama3}, exhibit strong reasoning and contextual understanding abilities, motivating their integration with diffusion models for improved text interpretation and image generation.
Liu \etal~\cite{liu2025llm4gen} proposed LLM4GEN, which combines LLM and CLIP representations to better capture complex prompt semantics.
Wu \etal~\cite{wu2025omnigen2} introduced OmniGen2, which uses a multimodal LLMs (MLLMs) as an encoder and adopts a fully decoupled dual-path architecture to achieve multimodal generation.
Liao \etal~\cite{liao2025imagegen} introduced ImageGen-CoT, which fine-tunes MLLMs using automatically collected datasets to improve contextual reasoning for T2I task.
Meanwhile, agent-based systems such as GenArtist~\cite{wang2024genartist} and MCCD~\cite{li2025mccd} leverage MLLMs for task decomposition, scene planning, and self-correcting generation.

In summary, although T2I models have made significant progress in semantic fidelity and visual quality~\cite{esser2024scaling, batifol2025flux}, they remain largely emotion-unaware.
This gap motivates generation methods to explicitly control affective expression.

\subsection{Alignment in Text-to-Image Generation}
Alignment in text-to-image generation aims to fine-tune the generative model to make the generated images more consistent with human preferences.
Early work largely focused on datasets.
Some methods~\cite{rombach2022high, podell2023sdxl} used aesthetic classifiers~\cite{schuhmann2022laionaesthetics} to filter data for fine-tuning.
Others~\cite{betker2023improving, segalis2023picture} improved textual supervision by rewriting image captions to enhance generative quality.
In addition, several works~\cite{kirstain2023pick, xu2023imagereward, wu2023human} constructed generative preference datasets to fine-tune generation models for better alignment with human preferences.
However, these approaches largely rely on dataset quality.

To reduce such dependence, reinforcement learning (RL) has been introduced to directly optimize generative models with external reward.
DDPO proposed by Black \etal~\cite{black2023training} and DPOK proposed by Fan \etal~\cite{fan2023dpok} fine-tune diffusion models to maximize learned preference rewards.
B$^2$-DiffuRL~\cite{hu2025towards} alleviates sparse-reward issues through backward progressive training and branch-based sampling.
Flow-GRPO~\cite{liu2025flow} is the first to integrate online policy gradient RL into flow matching models.
DanceGRPO~\cite{xue2025dancegrpo} further adapts Group Relative Policy Optimization (GRPO) to visual generation,
while MixGRPO~\cite{li2025mixgrpo} introduces a sliding window mechanism to reduce optimization overhead and accelerate convergence.
Despite these advances, existing alignment methods mostly rely on general-purpose reward functions such as PickScore and ImageReward, with limited consideration of emotion in reward design.

\subsection{Affective Image Synthesis}
Affective image synthesis (AIS) aims to generate or modify images to evoke specific emotions in viewers and has attracted growing attention in recent years.
Early work primarily use color transfer \cite{yang2008automatic, peng2015mixed, liu2018emotional, zhu2023emotional} or style transfer\cite{fu2022language, sun2023msnet, weng2023affective} to elicit emotions.
However, these methods operate on low-level color and texture cues without modeling high-level semantics that are essential for conveying emotions~\cite{zhao2023toward, zhao2024err}.
Inspired by psychological findings that visual content strongly influences emotional perception~\cite{brosch2010perception}, Yang \etal~\cite{yang2024emogen} proposed EmoGen, which aligns an emotion space with CLIP semantics to enhance the affective expressiveness of generated images.
Furthermore, Yang \etal~\cite{yang2025emoedit} introduced EmoEdit, which trains an Emotion Adapter on EmoSet~\cite{yang2023emoset} to modify image content toward target emotions, marking an early attempt at affective image manipulation.
More recently, Dang \etal~\cite{dang2025emoticrafter} proposed EmotiCrafter, which models continuous emotion dimensions (\eg, valence–arousal) within a diffusion framework to generate emotionally rich and nuanced images.
Yuan \etal~\cite{yuan2025coemogen} proposed CoEmoGen, which designs a hierarchical low-rank adaptation module to reduce the dependence on word-level attribute labels for emotion guidance.
Existing methods enhance emotion awareness but lack fine-grained content control.
To address this limitation, we introduce Controllable Emotional Image Content Generation (C-EICG) and propose EmoCtrl, which enables simultaneous control of content and emotion.

\section{Method}
\label{sec:method}

\begin{figure*}
	\centering
	\includegraphics[width=\linewidth]{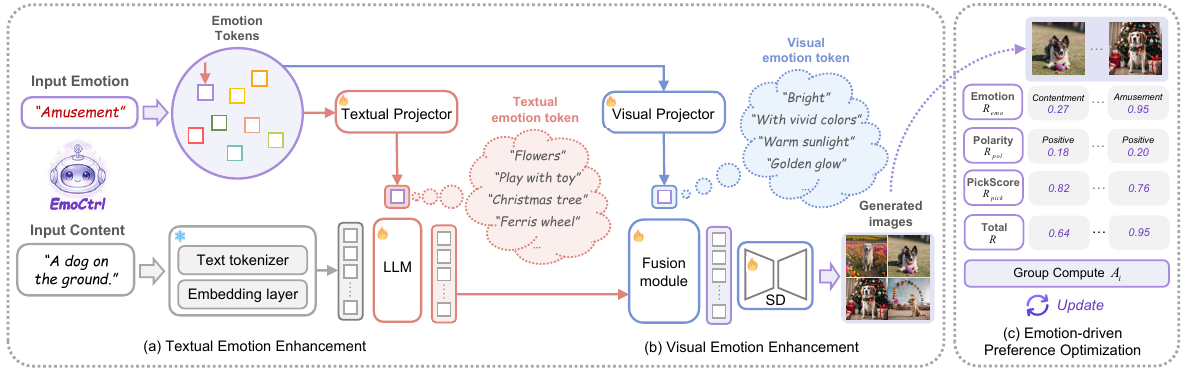}
	\vspace{-15pt}
	\caption{Overview of EmoCtrl. (a) Textual Emotion Enhancement: Emotion tokens are fused with content text in the LLM to enhance emotion at the semantic level. (b) Visual Emotion Enhancement: Emotion tokens are fused with the textual emotional condition within Stable Diffusion to produce affective images through visual cues. (c) Emotion-driven Preference Optimization: Refines the model with emotion-driven reward for better human alignment.}
	\Description{}
	\label{fig:method_2}
	\vspace{-10pt}
\end{figure*}

\subsection{Data Construction}
Existing AIS datasets~\cite{yang2023emoset, yang2025emoedit} focus on emotion labels but lack content conditions, while T2I datasets provide only semantic descriptions.
To support C-EICG, which requires both emotion and content control, we construct datasets tailored to this setting.
As content labeling is less ambiguous than emotion labeling, we extend existing AIS datasets with content annotations, as illustrated in Fig.~\ref{fig:method_1}.
EmoSet~\cite{yang2023emoset} is a large-scale dataset with each image labeled with an emotion category, while EmoEditSet~\cite{yang2025emoedit} contains image pairs, each annotated with an emotion category.
For EmoSet, we employed GPT-4V~\cite{achiam2023gpt} to generate detailed captions for each image and used LLaMA~\cite{grattafiori2024llama3} to extract visual concepts from the verified captions.
Since emotion and content serve as two orthogonal conditions, we filter out concepts with obvious emotional intent (\eg, Christmas tree, ghost) and retain the neutral ones (\eg, dog, building).
Consequently, annotators manually verified all generated captions and concepts, with low-agreement annotations re-extracted and re-verified.
For EmoEditSet, we used Qwen2.5-VL-72B~\cite{bai2025qwen2} to generate captions for each image and applied CLIP~\cite{radford2021clip} similarity filtering to remove misaligned caption–image pairs.
More implementation details are provided in the supplementary materials.
Eventually, we constructed two types of quadruplets: \{emotion, concept, caption, image\} for EmoSet+ and \{emotion, caption1, caption2, image\} for EmoEditSet+, where the input content is presented by either a concept or a sentence.
In total, we obtained 158,222 quadruplets, providing comprehensive supervision for C-EICG.

\subsection{EmoCtrl}
Based on the tailored datasets, \ie, EmoSet+ and EmoEditSet+, we propose EmoCtrl to generate affective images that align with the given content condition.
Emotions are complex and can be elicited by multiple factors~\cite{brosch2010perception}.
Inspired by prior studies in visual emotion analysis~\cite{yang2021stimuli}, we enhance affective expression from two complementary perspectives: textual descriptions (Sec.~\ref{sec:text_method}) and visual cues (Sec.~\ref{sec:visual_method}).
Furthermore, we introduce Emotion-driven Preference Optimization (Sec.~\ref{sec:preference_method}), which refines the generation policy using a carefully designed reward that jointly considers target emotion alignment, polarity consistency, and human preference.

\subsubsection{Textual Emotion Enhancement}
\label{sec:text_method}
Emotions are inherently abstract, yet in visual generation they are often evoked through concrete semantic concepts rather than explicit affective labels alone.
For example, a soccer game may evoke \textit{excitement}, while ghosts may induce \textit{fear}.
To bridge this gap, we enrich the LLM with \{emotion label, content label, affective prompt\} triplets from EmoSet+ and EmoEditSet+, enhancing its understanding of visual emotions.

As shown in Fig.~\ref{fig:method_2}~(a), we introduce Textual Emotion Enhancement, a module that explicitly injects emotional knowledge into the textual domain.
Specifically, inspired by textual inversion~\cite{gal2022ti}, we maintain a set of emotion tokens $V = \{v^0, v^1, \dots, v^7\} \in \mathbb{R}^{d}$, $d=1024$, where each token $v^k$ acts as an anchor capturing the intrinsic properties of a specific emotion category (\eg, amusement, fear).
Given an input content, we first tokenize it into a sequence of tokens $T = \{t_1, t_2, \dots, t_M\}$ and obtain its content embeddings $f_t = E_{\text{t}}(T)\in \mathbb{R}^{M \times d_1}$, $d_1=1024$, through the text embedding layer.
Meanwhile, the target emotion token is fed into a textual projector to obtain a textual emotion token $v_{t}^k \in \mathbb{R}^{d_1}$, which maps the emotion-specific representation into the language space of the LLM.
The visualization results of the textual emotion token further shows their effectiveness in Sec.~\ref{sec:experiments}.
The textual emotion token is then injected into the LLM together with the content embeddings, enabling the model to produce an emotion-enhanced textual condition that remains semantically consistent with the input content.
To better align the textual emotion token with the LLM, we adopt LoRA~\cite{hu2022lora} fine-tuning, which allows the model to learn how to effectively integrate the emotion token with content tokens, thereby enhancing the affective expression in the generated captions.
Given the ground-truth affective caption $q$, we optimize the module using the standard auto-regressive cross-entropy objective:
\begin{align}
	\label{eq:llmloss}
	\mathcal{L}_t = -\sum_{i=1}^{M} \log P(q_i \mid q_{<i} , v_t^k, \theta),
\end{align}
where $M$ is the length of $q$ and $\theta$ denotes the trainable parameters, including the learnable emotion tokens, the textual projector, and the LoRA parameters.

\subsubsection{Visual Emotion Enhancement}
\label{sec:visual_method}
While Textual Emotion Enhancement enriches the textual condition with emotion-relevant semantics, emotional perception in images is also strongly influenced by perceptial factors such as color, brightness, and style~\cite{yang2021stimuli}.
To further enhance affective expression at the visual level, we introduce Visual Emotion Enhancement module that injects emotion-aware perceptual cues into the image generation process.

Different from the textual branch, which maps the shared emotion token into the language space of the LLM, the visual branch projects the same emotion token into the conditioning space of the diffusion model.
Specifically, given the shared emotion token set $V = \{v^0, v^1, \dots, v^7\} \in \mathbb{R}^{d}$, for a target emotion category $k$, we select the corresponding emotion token $v^k$ and feed it into visual projector to obtain a visual emotion token $v_{v}^k \in \mathbb{R}^{d_2}$, $d_2=2048$.
The visual emotion token encodes the perceptual characteristics associated with the target emotion in the visual generation space.

Given the emotion-enhanced textual condition generated by LLM, we further encode it with the text encoder to obtain in textual features $f_v = E_{\text{vis}}(c_t)\in \mathbb{R}^{N \times d_2}$, where $c_t$ denotes the textual condition produced by the Textual Emotion Enhancement module.

To jointly exploit semantic and perceptual affective cues, we introduce a fusion module that integrates the textual features and the visual emotion token into a unified emotion-aware condition:
\begin{equation}
	c_v = \mathcal{F}(f_v, v_{v}^k),
\end{equation}
where $\mathcal{F}$ denotes the fusion module and $c_v$ is the final condition for image generation.
In this way, the semantic cues inferred by the LLM and the perceptual cues derived from the shared emotion token can complement each other, enabling the model to better express the target emotion while preserving the input content.
Finally, the diffusion model is optimized with the standard denoising objective:
\begin{equation}
	\label{eq:diffloss}
	\mathcal{L}_v = \mathbb{E}_{z_0, \epsilon, t, c_v} \big[ \|\epsilon - \varepsilon_\theta(z_t, t, c_v)\|_2^2 \big],
\end{equation}
where $\epsilon$ represents the added noise, $\epsilon_\theta$ denotes the denoising network and $z_t$ indicates the latent noised to time $t$.

\begin{figure*}
	\centering
	\includegraphics[width=0.95\linewidth]{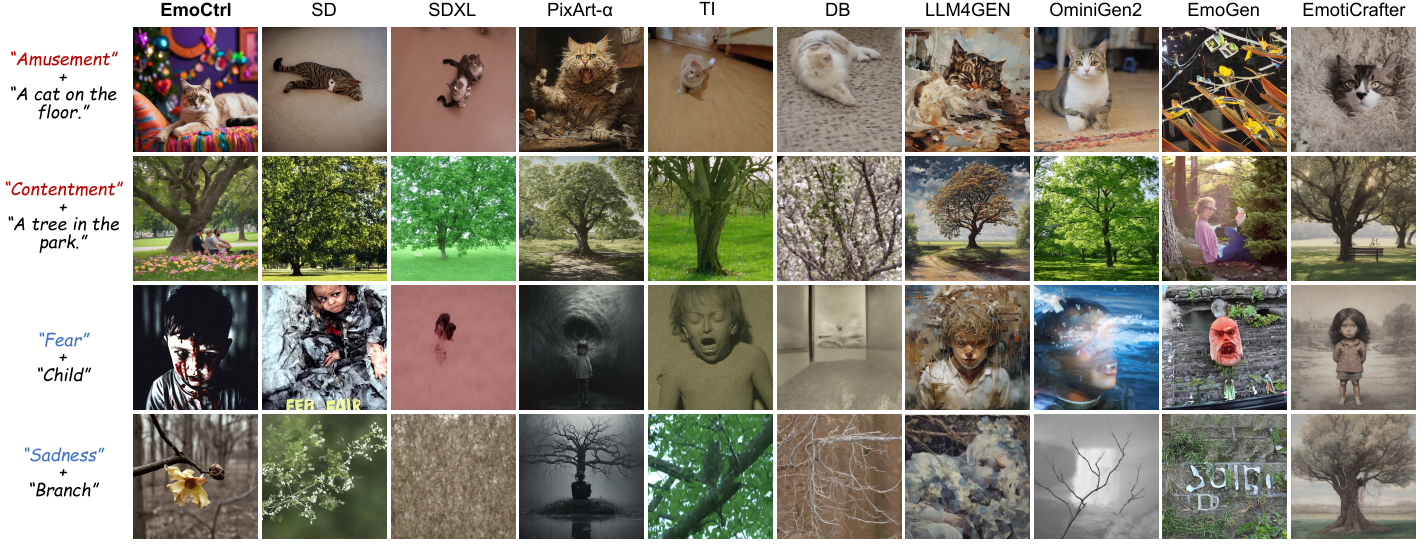}
	\vspace{-5pt}
	\caption{Comparison with state-of-the-art methods, showing EmoCtrl is superior in both content and emotion aspects.}
	\Description{}
	\label{fig:sota}
	\vspace{-10pt}
\end{figure*}

\subsubsection{Emotion-driven Preference Optimization}
\label{sec:preference_method}
While textual and visual emotion enhancement modules enrich EmoCtrl with affective cues, they do not consider whether the generated images align with human preference.
We therefore introduce Emotion-driven Preference Optimization to further refine EmoCtrl with reward signals tailored to emotion quality and human preference.
Specifically, we treat the diffusion model as a conditional generation policy and assign a composite reward to each generated image.

\paragraph{Emotion-driven Reward.}
PickScore~\cite{kirstain2023pick} is widely used for human preference alignment, as it provides a general signal of human judgment.
However, it mainly captures overall visual quality and text-image alignment, rather than emotion accuracy and intensity. 
Therefore, EmoCtrl further introduces explicit emotion-aware supervision at both the category and polarity levels during preference optimization.
Inspired by prior work~\cite{9524517}, progressive rewards can offer dense supervision for policy optimization.
Accordingly, we formulate a continuous emotion reward using a pretrained emotion classifier. Following Mikels' emotion model~\cite{mikels2005emotional}, we consider eight discrete emotions and further group them into positive (amusement, awe, contentment, excitement) and negative (anger, disgust, fear, sadness) polarities. 
Given a generated image $x$ and target emotion $e_t$, the classifier outputs a probability distribution $p (e_t | x)\in \mathbb{R}^8$ over the emotion classes via softmax.

We first define an emotion reward to measure the confidence of the target emotion:
\begin{equation}
	R_{\text{emo}}(x) = p (e_t | x),
\end{equation}
where $p (e_t | x)$ denotes the predicted probability of input image $x$ on target emotion $e_t$.

Since optimizing only the target category may lead to sparse supervision, especially in the early stages of optimization, we further introduce a polarity reward.
Specifically, we define two mutually exclusive sets according to the polarity of the target emotion $e_t$: the congruent polarity set $\mathcal{E}_{\text{same}}$ and the conflicting polarity set $\mathcal{E}_{\text{opp}}$. 
The polarity reward is computed as
\begin{align}
	R_{\text{pol}}(x) = \alpha \!\!\!\!\!\sum_{e_i \in \mathcal{E}_{\text{same}}}\!\!\!\!\! p (e_i | x) - \beta \!\!\!\!\!\sum_{e_j \in \mathcal{E}_{\text{opp}}}\!\!\!\!\!p (e_j | x),
\end{align}
where $\alpha$ and $\beta$ are weighting factors.

Final reward jointly consider general human preference and emotion-specific alignment, encouraging the model to generate images that are both visually preferred and emotionally faithful:
\begin{equation}
	\label{eq:reward}
	R(x) = \omega_1 R_{\text{emo}}(x) + \omega_2 R_{\text{pol}}(x) + \omega_3 R_{\text{pick}}(x, c_t),
\end{equation}
where $\omega_1$, $\omega_2$, and $\omega_3$ are weighting factors,  $R_{\text{pick}}(x, c_t)$ denotes PickScore between the given image $x$ and the content condition $c_t$.

\paragraph{GRPO Optimization.}
To exploit relative reward differences among grouped samples while maintaining stable policy updates, we adopt Group-Relative Policy Optimization (GRPO)~\cite{shao2024deepseekmath}.
For each condition $c_t$, we sample a group of $G$ images $\{x_i\}_{i=1}^G$ and compute their corresponding rewards $\{R(x_i)\}_{i=1}^G$.
GRPO normalizes the rewards within each group to calculate relative advantages:
\begin{equation}
	\label{eq:advantage}
	A_i = \frac{R(x_i) - \mu_R}{\sigma_R},
\end{equation}
where $\mu_R$ and $\sigma_R$ denote the mean and standard deviation of rewards within the group.
This relative advantage mechanism encourages the model to favor better-performing samples based on within-group comparison, without being overly sensitive to absolute reward scale.
The overall objective is formulated as
\begin{align}
	\mathcal L_{\text{GRPO}}
	& = -\,\mathbb E_{x_i \sim \pi_{\theta_{\text{old}}}}
	\Big[
	\min\!\Big(
	r_i(\theta)A_i,\;
	\operatorname{clip}_{[1-\epsilon,\,1+\epsilon]}\!\big(r_i(\theta)\big)A_i
	\Big)
	\Big] \nonumber                                                                  \\
	& \quad + \lambda\,D_{\mathrm{KL}}\!\big(\pi_\theta \,\|\, \pi_{\text{ref}}\big),
\end{align}
where $r_i(\theta) = \frac{\pi_\theta(x_i \mid c_t)}{\pi_{\theta_{\text{old}}}(x_i \mid c_t)}$ is the probability ratio,
and $\lambda$ controls the strength of the Kullback-Leibler (KL) penalty.
This penalty prevents the exploratory policy $\pi_\theta$ from deviating excessively from the pretrained base model $\pi_{\text{ref}}$, thereby preserving the model's generative capabilities.
\section{Experiments}
\label{sec:experiments}

\subsection{Dataset and Evaluation}

\paragraph{Dataset}
EmoCtrl is trained on the constructed EmoSet+ and EmoEditSet+ datasets.
For fair comparison, we construct a inference set containing 76 content descriptions, including 46 concepts (\eg, ``Beach'', ``Couple'') and 30 captions (\eg, ``A person is walking on a street'', ``A bird is flying in the sky'').
Each description is paired with 8 emotion categories, resulting in 608 images in total.
To ensure a data distribution consistent with the training set, the 46 concepts are derived from EmoSet+, while the captions are generated by GPT-5 based on these concepts.
For more details, please refer to the supplementary materials.

\paragraph{Evaluation Metrics}

As our task targets both content alignment and emotion expressiveness, we first employ corresponding metrics to evaluate these two aspects.
For emotion fidelity, following~\cite{yang2024emogen}, we adopt Emotion Accuracy (Emo-A) to measure whether the generated image aligns with the target emotion using a pre-trained classifier on EmoSet~\cite{yang2023emoset}.
Since our goal is to inject emotional cues into given content, the conventional CLIP score~\cite{radford2021clip} is not a fair measure for content alignment.
We therefore propose CLIP Accuracy (CLIP-A) to evaluate how well the generated image preserves the input content.
To jointly assess both objectives, we further introduce Emotion–Content Joint Accuracy (EC-A), which validates whether a generated result satisfies both emotion and content conditions.
For more details, please refer to the supplementary materials.
Additionally, we use LPIPS~\cite{zhang2018lpips} to evaluate low-level perceptual diversity and Semantic Clarity (Sem-C)~\cite{yang2024emogen} to assess high-level recognizability.

\subsection{Comparisons}

We compare EmoCtrl with the most relevant and state-of-the-art methods across four categories:
(1) Foundation models: Stable diffusion \cite{rombach2022high}, SDXL \cite{podell2023sdxl} and PixArt-$\alpha$ \cite{chen2023pixart};
(2) Personalized generation: Textual inversion (TI) \cite{gal2022ti} and Dreambooth (DB) \cite{ruiz2023dreambooth};
(3) Language Model-enhanced: LLM4GEN \cite{liu2025llm4gen} and OmniGen2 \cite{wu2025omnigen2};
(4) Affective Image Synthesis: EmoGen \cite{yang2024emogen}, EmotiCrafter \cite{dang2025emoticrafter}.

\paragraph{Qualitative Comparisons}

We present qualitative comparisons in Fig.~\ref{fig:sota}.
Foundation models (SD, SDXL, and PixArt-$\alpha$) exhibit limited emotional understanding: they can depict the given content but often lack affective awareness, sometimes leading to distorted or emotionless results.
Personalized generation methods (TI and DB) tend to learn abstract emotions as styles, resulting in rigid or repetitive visual patterns.
Language model-based methods (LLM4GEN and OmniGen2) incorporate partial emotional cues through semantic expansion but frequently produce ambiguous or semantically inconsistent results, failing to effectively combine emotion and content cues.
Among emotion-related methods, EmoGen often generates images with unclear semantics and overfitted textures, while EmotiCrafter applies vintage-like filters that blur fine details.
In contrast, EmoCtrl consistently produces images that are both emotion-expressive and content-faithful.
Benefiting from textual and visual emotion enhancement, our results are semantically explicit and perceptually expressive.

\begin{table}[htbp]
	\centering
	\footnotesize
	\caption{
		Comparisons with four categories of state-of-the-art methods across five evaluation metrics.
	}
		\vspace{-5pt}
		\begin{tabular}{lccccc}
			\toprule
			Method                                   & Emo-A↑              & CLIP-A↑             & EC-A↑            & LPIPS↑         & Sem-C↑         \\
			\midrule
			SD-1.5 \cite{rombach2022high}            & 16.12\%             & 78.95\%             & 13.65\%          & 0.672          & \underline{0.606}          \\
			SDXL \cite{podell2023sdxl}               & 22.37\%             & 75.16\%             & 16.28\%          & 0.563          & 0.535          \\
			PixArt-$\alpha$ \cite{chen2023pixart}    & 20.72\%             & \underline{85.69\%} & 17.76\%          & 0.652          & 0.590          \\
			\midrule
			TI \cite{gal2022ti}                      & 38.82\%             & 66.94\%             & 22.53\%          & 0.537          & 0.525          \\
			DB\cite{ruiz2023dreambooth}              & 33.75\%             & 81.53\%             & \underline{24.86\%}          & 0.588          & 0.550          \\
			\midrule
			LLM4GEN \cite{liu2025llm4gen}            & 21.22\%             & 74.51\%             & 15.46\%          & 0.557          & 0.506          \\
			OmniGen2 \cite{wu2025omnigen2}           & 25.00\%             & \textbf{89.97\%}    & 21.22\%          & \textbf{0.708} & 0.595          \\
			\midrule
			EmoGen \cite{yang2024emogen}             & \underline{45.23\%} & 43.42\%             & 14.97\%          & \underline{0.701}          & 0.539          \\
			EmotiCrafter \cite{dang2025emoticrafter} & 24.67\%             & 82.73\%             & 20.23\%          & 0.485          & 0.563          \\
			\midrule
			\textbf{EmoCtrl}                         & \textbf{64.64\%}    & 83.06\%             & \textbf{50.99\%} & 0.699          & \textbf{0.673} \\
			\bottomrule
		\end{tabular}
	\vspace{-0.05in}
	\vspace{-5pt}
	\label{tab:table1}
\end{table}

\paragraph{Quantitative Comparisons}

As shown in Table~\ref{tab:table1}, EmoCtrl achieves the best overall performance across most metrics.
Specifically, it obtains the highest Emo-A (64.64\%), indicating its strong capability to generate emotional elements aligned with the target emotion.
For CLIP-A, EmoCtrl achieves comparable results to text-to-image models, as the introduction of emotion-related cues may slightly affect strict content matching.
Despite the dual-target challenge of C-EICG, EmoCtrl attains the highest EC-A (50.99\%), substantially outperforming the second-best method DB (24.86\%), highlighting its balanced performance on both emotion and content alignment.
In terms of perceptual quality, our method achieved a moderately high LPIPS score (0.699), demonstrating a certain degree of competitiveness.
Regarding semantic consistency, Sem-C also reaches the highest value (0.673), consistent with the qualitative results in Fig.~\ref{fig:sota}.
Overall, these quantitative results demonstrate that EmoCtrl effectively balances emotion expression and content fidelity, outperforming existing state-of-the-art methods in emotion-aware text-to-image generation.

\begin{table}[h]
	\centering
	\footnotesize
	\caption{
		User preference study. The numbers indicate the percentage of participants who vote for the result.
	}
	\vspace{-5pt}
		\begin{tabular}{lccc}
			\toprule
			Method                       & Emotion evoking ↑           & Content fidelity ↑           & Balance ↑                   \\
			\midrule
			SDXL \cite{podell2023sdxl}   & 0.94 $\pm$ 2.82\%           & 1.35 $\pm$ 4.56\%            & 1.15 $\pm$ 3.60\%           \\
			TI \cite{gal2022ti}          & 5.05 $\pm$ 4.19\%           & 6.46 $\pm$ 5.75\%            & 5.76 $\pm$ 4.41\%           \\
			EmoGen \cite{yang2024emogen} & 5.21 $\pm$ 4.53\%           & 5.42 $\pm$ 4.34\%            & 5.31 $\pm$ 4.20\%           \\
			\textbf{EmoCtrl}                         & \textbf{88.75 $\pm$ 8.88\%} & \textbf{86.77 $\pm$ 11.81\%} & \textbf{87.76 $\pm$ 9.76\%} \\
			\bottomrule
		\end{tabular}
	\vspace{-0.05in}
	\vspace{-5pt}
	\label{tab:table2}
\end{table}

\paragraph{User Study}

We further conduct a user study to evaluate human preference.
A total of 48 participants with diverse ages and backgrounds were invited, and each session lasted about 20 minutes.
The study included 40 triplets of emotion–content conditions, each paired with four generated images from different methods: SDXL, TI, EmoGen, and EmoCtrl.
Participants evaluated the results along two dimensions: semantic fidelity and emotional evocation.
As shown in Table~\ref{tab:table2}, EmoCtrl received the highest number of votes across all criteria.
Despite the inherent challenge of maintaining semantic fidelity while evoking emotion, EmoCtrl achieved top preferences in both emotion (88.75\%) and content (86.77\%), and a balanced score of 87.76\%.
Compared with the quantitative results in Table~\ref{tab:table1}, the user study further confirms that EmoCtrl is more favored by humans than by conventional metrics, validating its effectiveness in both emotion fidelity and content consistency.

\begin{table}[htbp]
	\centering
	\footnotesize
	\caption{
		Ablation of EmoCtrl. Baseline is the original Stable Diffusion. $v_t$ and $v_v$ denote the textual and visual emotion tokens, respectively, while EDPO stands for Emotion-driven Preference Optimization.
	}
		\vspace{-5pt}
		\begin{tabular}{lccccc}
			\toprule
			Setting & Emo-A↑ & CLIP-A↑ & EC-A↑ & LPIPS↑ & Sem-C↑            \\
			\midrule
			baseline & 12.50\% & \underline{96.05\%} & 12.01\% & 0.692 & 0.633 \\
			w/o $v_t$ & 12.99\% & \textbf{96.88\%} & 12.50\% & 0.679 & 0.670 \\
			w/o $v_v$ & \textbf{65.30\% } & 82.24\% & 50.33\% & 0.693 & \underline{0.672} \\
			w/o EDPO & \underline{64.64\%} & 80.92\% & \underline{49.34\%} & \textbf{0.703} & 0.658 \\
			\textbf{EmoCtrl} & \underline{64.64\%} & 83.06\% & \textbf{50.99\%} & \underline{0.699} & \textbf{0.673} \\
			\bottomrule
		\end{tabular}
	\vspace{-0.05in}
	\vspace{-5pt}
	\label{tab:ablation}
\end{table}

\subsection{Ablation Study}

We conduct ablation studies to examine the contributions of the textual emotion token $v_t$, the visual emotion token $v_v$, and Emotion-driven Preference Optimization (EDPO).
As shown in Table~\ref{tab:ablation}, removing any component degrades the overall performance of EmoCtrl, illustrating the difficulty of the C-EICG task.
In particular, removing $v_t$ leads to lower emotion-related performance, indicating that textual emotion enhancement is important for grounding target emotions in appropriate high-level semantics.
Removing $v_v$ causes the largest performance drop, suggesting that visual emotion enhancement is crucial for translating emotion semantics into effective perceptual cues.
Moreover, removing EDPO consistently weakens the results, showing that preference optimization further improves emotional alignment beyond supervised affective conditioning.

\begin{figure}
	\centering
	\includegraphics[width=\linewidth]{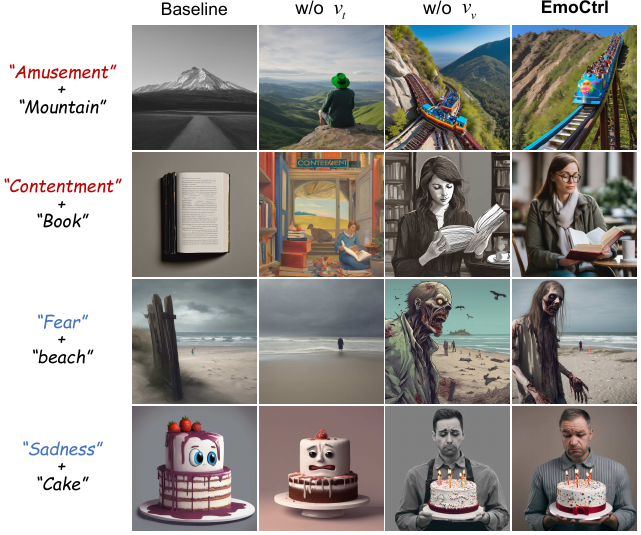}
	\vspace{-15pt}
	\caption{Ablation study of EmoCtrl, showing the contributions of textual emotion tokens ($v_t$) and visual emotion tokens ($v_v$).}
	\Description{}
	\label{fig:ablation}
	\vspace{-10pt}
\end{figure}

The qualitative comparisons in Fig.~\ref{fig:ablation} further illustrate the complementary roles of $v_t$ and $v_v$.
For example, given the condition ``sadness + cake'', the baseline model tends to rely on anthropomorphic cues, directly generating a crying personified cake rather than conveying sadness through realistic scene composition or visual atmosphere.
A similar phenomenon can be observed without $v_t$, indicating that textual emotion tokens are important for steering the model toward emotion-relevant semantics.
When $v_v$ is removed, the model is less prone to this anthropomorphic tendency because $v_t$ provides emotional semantic guidance.
However, the generated results become less realistic, often appearing cartoonish or black-and-white.
By jointly using $v_t$, $v_v$, and EDPO, EmoCtrl achieves the best balance between content fidelity and emotional expressiveness.

\begin{figure*}
	\centering
	\includegraphics[width=0.98\linewidth]{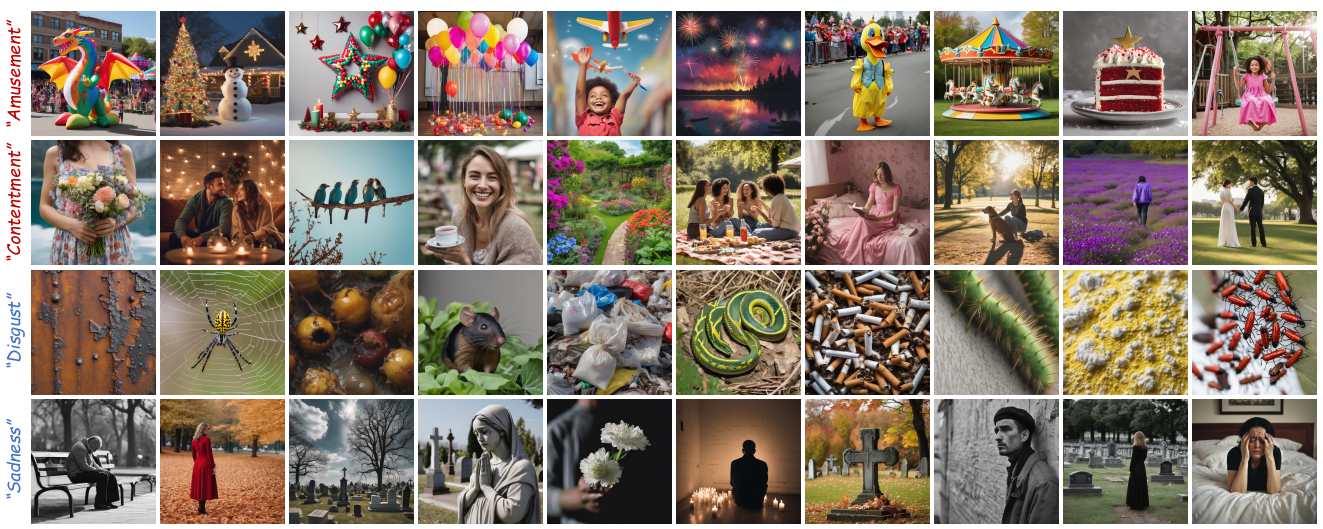}
	\vspace{-5pt}
	\caption{
		Visualization of textual emotion tokens. Each row shows images generated from a specific emotion, while each column reflects diverse semantic expressions. The results show that EmoCtrl produces rich content diversity while maintaining consistent emotional tone.
	}
	\Description{}
	\label{fig:vis}
\end{figure*}

\begin{figure*}
	\centering
	\includegraphics[width=0.98\linewidth]{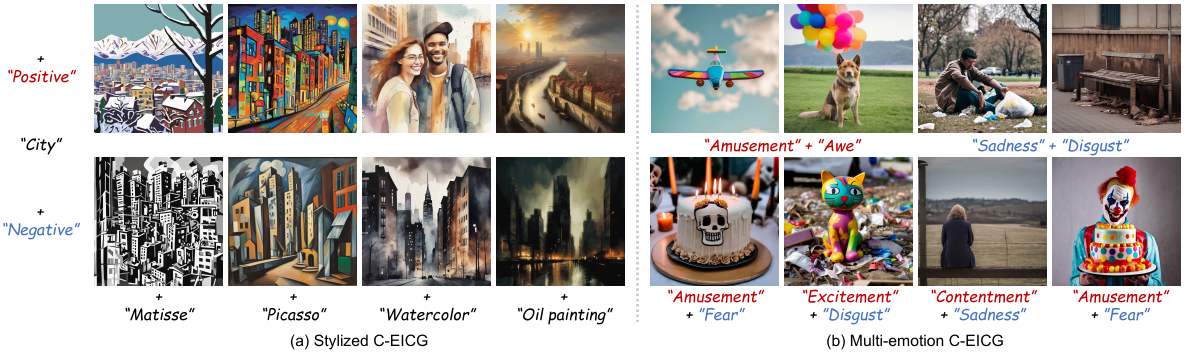}
	\caption{EmoCtrl can be applied to creative art generation, including (a) style control and (b) multi-emotion condition.}
	\Description{}
	\label{fig:app}
	\vspace{-10pt}
\end{figure*}

\subsection{Visualization}
\label{sec:vis}

We visualize the generative behavior of textual emotion tokens in Fig.~\ref{fig:vis}.
Each row corresponds to one emotion, and each column shows diverse outputs generated solely from that token.
Without any content prompts, EmoCtrl produces a wide range of affective semantics, indicating that the tokens effectively steer EmoCtrl toward emotion-aware generation.
Amusement and contentment yield bright and lively objects and scenes, like ballon, picnic and garden.
While disgust produces unpleasant textures and repulsive subjects, sadness consistently generates muted tones and somber atmospheres.
These results demonstrate that each textual emotion token embeds a strong emotional prior that generalizes across varied content.

\subsection{Applications}

\paragraph{Stylized C-EICG}
As shown in Fig.~\ref{fig:app} (a), EmoCtrl can generate creative results conditioned on emotion, artistic style, and semantic content.
Given the same content ``City'', the columns display four artistic styles, while the rows apply either a positive or negative emotion.
Positive emotions produces bright and vivid cityscapes featuring affective cues such as snow-capped mountains, neon lights, affectionate couples, and sunrises.
In contrast, negative emotion yields darker and more somber interpretations with elements like moons, shadows, and mist.
These results demonstrate that EmoCtrl can reliably modulate emotional tone while preserving both content semantics and artistic style.

\paragraph{Multi-emotion C-EICG}

To evaluate the robustness and independence of textual and visual emotion tokens, we mix different emotions during generation.
In Fig.~\ref{fig:app}~(b), the first row combines emotions with the same polarity, enhancing the affect in a consistent direction (\eg, Amusement+Awe for a toy airplane or Sadness+Disgust for a shabby bench).
The second row mixes opposite polarities.
The first two cases show conflicting emotions tied to different semantics, while the latter two involve semantics that are naturally emotion-ambiguous, such as a man with a lonely figure or a clown.
\section{Conclusion}
\label{sec:conclusion}

\paragraph{Discussion}

We propose EmoCtrl, a method for generating images that evoke specified emotions from user-provided content, supported by a newly constructed training dataset.
EmoCtrl learns visual emotion semantics with only a small number of learnable parameters and performs emotion augmentation at both the textual and visual levels.
Extensive experiments show that our method achieves a strong balance between semantic fidelity and emotional expressiveness. User studies, visualizations, and ablation analyses further validate the effectiveness of EmoCtrl, while downstream applications demonstrate its ability to handle a wide range of emotion-related generative tasks.

\paragraph{Limitations}
Our method can only accept text as input.
In real-world scenarios, users may want to use images as input to achieve customized emotional image content generation.
Furthermore, the model's emotional knowledge largely relies on EmoSet, which may introduce potential bias due to the limited data.
As visual emotion evolves, expanding to high-quality, high-resolution visual emotion datasets will raise the performance ceiling of models.
Our method primarily enhances the image's ability to evoke specific emotions by adding semantics. Existing image-text metrics are mostly used to evaluate consistency and cannot effectively evalua
te our method, which can lead to bias and necessitates more human feedback.


\bibliographystyle{ACM-Reference-Format}
\bibliography{emoctrl}
\end{document}